\documentclass[letterpaper, 10 pt, conference]{ieeeconf}  

\IEEEoverridecommandlockouts                              

\usepackage{graphics} 
\usepackage{graphicx} 

\title{\LARGE \bf
Teleoperation System Using Past Image Records \\ Considering Narrow Communication Band
}

\author{Noritaka Sato, Masataka Ito and Fumitoshi Matsuno
\thanks{N. Sato is with department of Computer Science and Engineer, Nagoya Institute of Technology, Gokiso-cho, Showa-ku, Nagoya, Aichi 466-8555, Japan.
        {\tt\small sato.noritaka@nitech.ac.jp}}%
\thanks{M. Ito was with department of Mechanocal Engineering and Intelligent Systems, The University of Electro-Communications, Chofugaoka, Cofu, Tokyo, Japan.}
\thanks{F. Matsuno is with department of Mechanical Engineering and Science , Kyoto University, Kyotodaigaku-Katsura, Nishikyo-ku, Kyoto, 615-8540, Japan.}
}

\begin{document}

\maketitle
\thispagestyle{empty}
\pagestyle{empty}

\begin{abstract}

Teleoperation is necessary when the robot is applied to real missions, for
example surveillance, search and rescue.
We proposed teleoperation system using past image records (SPIR).
SPIR virtually generates the bird's-eye view image by
overlaying the CG model of the robot at the corresponding current position on
the background image which is captured from the camera mounted on the robot at a
past time.
The problem for SPIR is that the communication bandwidth is often narrow in some
teleoperation tasks.
In this case,
the candidates of background image of SPIR are few and the position of the robot
is often delayed.
In this study, we propose zoom function for insufficiency of candidates of
the background image and additional interpolation lines for the delay of the
position data of the robot.
To evaluate proposed system, an outdoor experiments are carried out.
The outdoor experiment is conducted on a training course of a driving school.
\end{abstract}

\section{INTRODUCTION}

\subsection{Background}

Recently, robots are used in order to complete tasks instead of humans
in the place where a human cannot enter.
Some of them are full autonomous,
but even in the case of an autonomous robot,
teleoperation is often necessary when the robot is applied to real missions, for
example surveillance, search and rescue.
\par
Generally, in the human-robot interface for the teleoperation,
images from a camera mounted on the robot are displayed.
However, if the camera is mounted on the front of the robot and its axis is
fixed to the forward direction, it is difficult for an operator to recognize the
width of the robot from the image and understand the relation between the robot
and the environment around the robot. Therefore, the operator has heavy workloads and a beginner cannot remote control the robot well.
In addition, in the teleoperation interface,
when the communication is unstable,
bad-quality images may be displayed and the images may have the delays,
because the information volume of an image is large.
When a robot runs on rough terrain, the images from the mounted camera on it are not stable because of robot attitude changes.
It is very hard for the operator to watch these nonsteady vibrately images for
remote control.
Therefore we should develop an interface to reduce the operator's workloads.
\par

\subsection{Related work}

To solve these drawbacks, some interfaces for teleoperation of a robot are
proposed.
Yanco et al. analyze the real teleoperation tasks and find situation awareness
is important to reduce workload of an operator \cite{Yanco}.
To improve situation awareness in the teleoperation, Yanco et al. recommends
providing the more spatial information to the operator. 
To display the robot and its surroundings,
Shiroma et al. attach a long pole on the robot whose top end has a camera
\cite{EI}.
Murphy et al. take vision supports from other robots' cameras where
the operated robot is on the image from a camera mounted on other robots
\cite{multi}. However these interfaces have disadvantages that the robot's size is increased or the mission often became slow because the operator should control two
or more robots. Nguyen et al. and Saitoh et al. proposed the interface in which
a virtual 3D environment including a CG model of the robot is displayed
\cite{VR, MVE}. And Nielsen et al. proposed the interface in which a CG model of
the robot and an image from the camera mounted on the robot are displayed on the environment map
\cite{Map}. However, these methods need 3D modeling or map building of the
environment and it often takes much time to generate them.
Therefore, these systems are difficult to implement to robots in the disaster area where the robot has error of self localization
and the communication is unstable.
\par
\begin{figure}[t]
\begin{center}
\includegraphics[width=0.8\hsize]{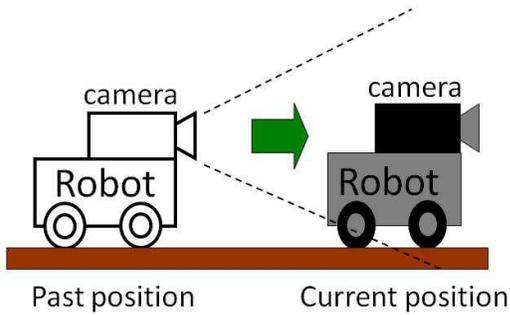}
\caption{Using a past image as a background image.}
\label{IM}
\end{center}
\end{figure}
\begin{figure}[t]
\begin{center}
\includegraphics[width=0.8\hsize]{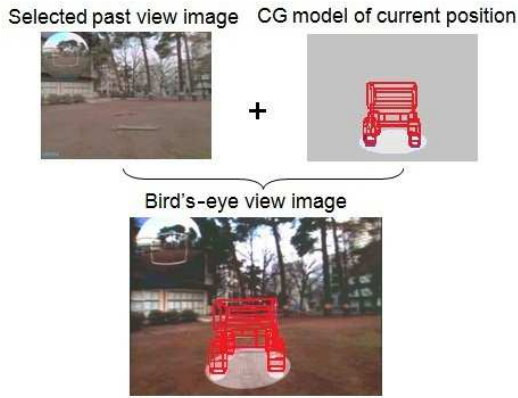}
\caption{Overview of SPIR.}
\label{overview}
\end{center}
\end{figure}

To develop the interface which is robust against bad communication conditions
and does not need heavy computational power and high cost sensors, Matsuno et
al. proposed an original idea of the teleoperation system using past image records (SPIR), which is
effective to an unknown environment \cite{patent}. This system uses an image
captured at a past time as a background image (Fig. \ref{IM}) and
overlays a CG model of the robot on it at the corresponding current position and orientation. In this way, the system virtually displays the
current robot situation and its surroundings from the past point of view
of the camera mounted on the robot (Fig. \ref{overview}).
The system can generate a bird's-eye view image which includes the teleoperated
robot and its surroundings,
and real time vibrations of the image is eliminated by using a fixed background
image captured in the past time.
\par

\subsection{Principle of SPIR}

 The algorithm of SPIR is following: \\
\\
(Step 1) Get robot's position and orientation\\
 The system gets current robot's position and orientation based on odometry,
 GPS, SLAM algorithm and so on. In SPIR, the error of the position is canceled when the background image
 is switched, because the position of the virtual robot (CG model) is depends
 on relative positions between current robot position and past robot position that the
 background image was captured.\\
 \\
(Step 2) Save the image and its position \\
The system stores images from the camera mounted on the robot on the buffer
(temporary memory) as candidates of the background image. At this time, the
system also stores the position of the robot as the camera position where the
image is captured. The set of the stored image and  position is called
"past image record." \\
\\
(Step 3) Select the background image \\
The system selects an optimal image as a background.
In this study, the system switches the background image if the distance
between the current robot position and the past robot position where the
background image was taken is larger than a threshold value.
\\
\\
(Step 4)Generate bird's-eye view scene \\
The system generates a bird's-eye view scene by overlaying a CG model of the
robot at the corresponding current position and orientation on the
background image selected in Step(3).
\\
\\
By iterating above algorithm, the system displays the bird's-eye view image
to the operator (Fig. \ref{overview}). Detail descriptions of the
implementation of SPIR are reported in existing papers \cite{kagotani2,sice}.

\subsection{Research purpose}

In this study, we focus on operability in the case of the narrow communication bandwidth.
Because SPIR uses discrete images, the load of transmission is
reduced compared with traditional system which the operator controls the
robot by using the images sending in the real time from the robot.
However, if the communication bandwidth is narrow,
stored images in the database of the candidate of the background image may be
too few. Therefore, the system may provide a feeling of strangeness for the operator
in the case of the narrow communication band, because the size of the CG model
is changed significantly when the background image is switched. The longer the distance from the position where the background image is captured is, the smaller the size of CG model is.
If the candidates of the background image are few, the distance may be long.
In this study, we proposed zoom function to overcome this
problem.
\par
Moreover, in the case of the narrow communication band,
transmission delay will occur.
In this case, since the current position data of the robot also have delay,
the position data of the CG model on the background image is not correct.
Therefore it is difficult for the operator to teleoperate the robot,
because the operator misses the current position of the robot.
In this study, we propose additional interpolation lines that the operator
can predict the robot position easily.
And the operator can use them as
indicators for generating driving control commands of the robot.
\par
In the section 2, we proposed zoom function and additional interpolation lines in SPIR for
improving the robustness against the communication delay. 
In the section 3, we show evaluation experiments
in the outdoor environments. 
The section 4 is conclusion.

\section{SPIR considering narrow communication band}

\subsection{Zoom function}

\begin{figure}[t]
 	\centering
 	\includegraphics[width=0.9\hsize]{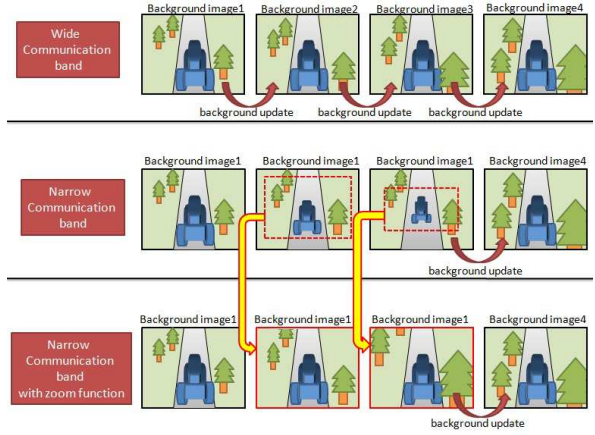}
	\caption{Snapshot images of SPIR with and without zoom fanction.}
	\label{fig:zoom}
\end{figure}

When the background image is changed, the relative relation between the current
position of the robot and the past viewpoint where the selected background image
was captured is discretely changed. If the distance of the past viewpoints
before and after the background image is updated is small, the system makes a
frequent small change of the size of the CG model of the robot on the monitor
when the background image is switched, as shown in the upper images of Fig. \ref{fig:zoom}.
In the case of the narrow communication band,
because many images cannot be sent, the relative distance of two positions
may be large.
The big change of the size of the CG model of the robot in the background image
as shown in the middle images of Fig. \ref{fig:zoom} is a source of the sense
of incongruity. To overcome this problem, a zoom function is
installed in SPIR in order to keep the size of CG model on the background
image for every sampling time, as shown in the bottom images of Fig.
\ref{fig:zoom}.

We define a vertical angle $\theta$ of the field of
view (FOV) as a parameter of a zooming ratio.
We explain about the calculation method of the vertical angle $\theta$ to keep
the ratio of CG models on the displayed background images.

\begin{figure}[t]
 	\centering
 	\includegraphics[width=0.99\hsize]{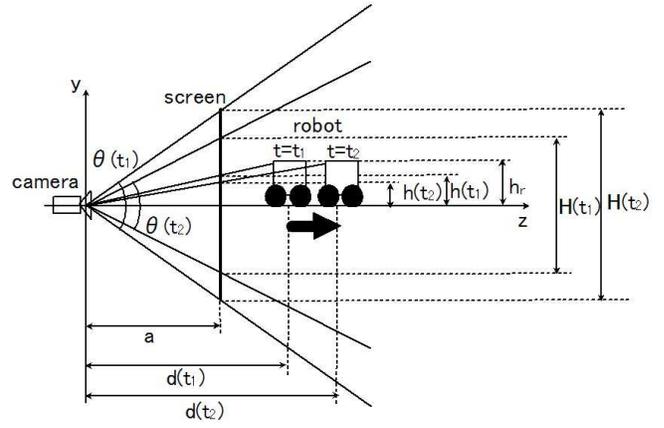}
 	\caption{Calculation of angle of field.}
 	\label{fig:calcAngle}
\end{figure}

As shown in Fig. \ref{fig:calcAngle}, $d(t)$ is the distance between the
viewpoint of the background image and the current position of the robot,
$\theta(t)$ is the vertical angle of FOV of the background image, $h$ is the
constant height of the robot, and $H(t)$ is the vertical length of FOV
positioned $d(t)$ away from the viewpoint. The system should keep
$h/H(t)$ to be a constant value $k (0<k<1)$ for every sampling,
because if this
ratio is kept, the size of the CG model on the background image is also not
changed regardless of the motion of the robot. From the geometric relationship,
\begin{equation}
d(t)\tan(\theta(t)/2)=H(t)/2, \ \ k=h/H(t).
\end{equation}
We obtain
\begin{equation}
\label{thetaZoom}
\theta(t)=2\tan^{-1}(h/2d(t)k).
\end{equation}


In the case that the background image does not change ($0 \le t < t_1$ in
Fig. \ref{fig:zoom}), the distance $d(t)$ changes continuously, then
$\theta(t)$ changes continuously based on Eq.(\ref{thetaZoom})to keep $h/H(t)$.
In the case that the background image is updated ($t=t_1$ in
Fig. \ref{fig:zoom}),
the distance $d(t)$ changes discretely, then $\theta(t)$ changes discretely to
keep $h/H(t)=k$.
By changing the angle $\theta(t)$ of the field of view to keep $k$
constant according to the Eq. (\ref{thetaZoom}), even if the background image is
switched, the size of the CG model on the background image is not changed.
\par
In proposed system, we used the OpenGL and OpenCL functions for image mapping
and image extraction.

\subsection{Additional line}

If the transmission delay occurs, the current position data of the robot
also has delay and the position of the CG model on the background image in
SPIR is not correct. Therefore it is difficult for the operator to understand
the current position of the robot and control the robot.
\par
To overcome this problem, we add lines on the displayed image generated by SPIR.
By displaying additional lines, the operator can easily predict the
coming position of the robot and control it.
\par
In this research, we introduce two types of additional lines;
\par
\begin{flushright}
\begin{figure}[t]
 	\centering
 	\includegraphics[width=0.8\hsize]{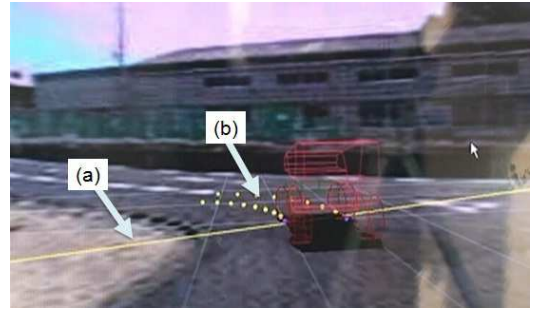}
	\caption{Additional lines.}
	\label{fig:additional_line}
\end{figure}
\end{flushright}

(a) Extended line of front wheel axis (Solid lines (a) in Fig.
\ref{fig:additional_line})\\ This line is an extension of the front wheel axis of the robot.
This line is overlaid on the background image according to the steering angle
of the robot. Because the operator can easily understand the center of the
rotation of the robot, the robot can smoothly turn by fixing this line on the
center of the corner.
\par
(b) Predictive trajectory (Dotted lines (b) in Fig.
\ref{fig:additional_line})\\
This line shows the predictive trajectory of the
wheel of the robot. By using the predictive trajectories,
an operator can easily estimate the motion of the robot.
By collimating this line with the edge of the course or the center line of the road,
the robot can run without swerving from the road.

\section{Evaluation experiment}

\subsection{Experimental method}

we valid the effectiveness of the zoom function and
the additional lines in the case of narrow communication band as explained in
the section 3. The number of subjects is 8, and we compared three methods:
(1)normal front camera whose angle of FOV is 60 degrees(Front Camera),
(2)existing SPIR without use zoom function and not add
lines (Existing SPIR)
and (3) extended SPIR with zoom function and add lines(Proposed SPIR2).
\par
\begin{flushright}
\begin{figure}[t]
 	\centering
 	\includegraphics[width=0.85\hsize]{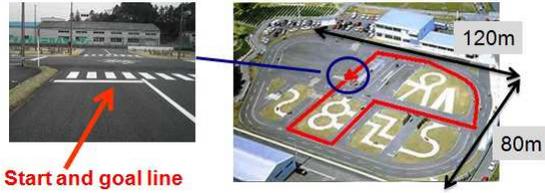}
	\caption{Course of examination.}
	\label{fig:course}
\end{figure}
\end{flushright}
\begin{figure}[t]
 	\centering
 	\includegraphics[width=0.55\linewidth]{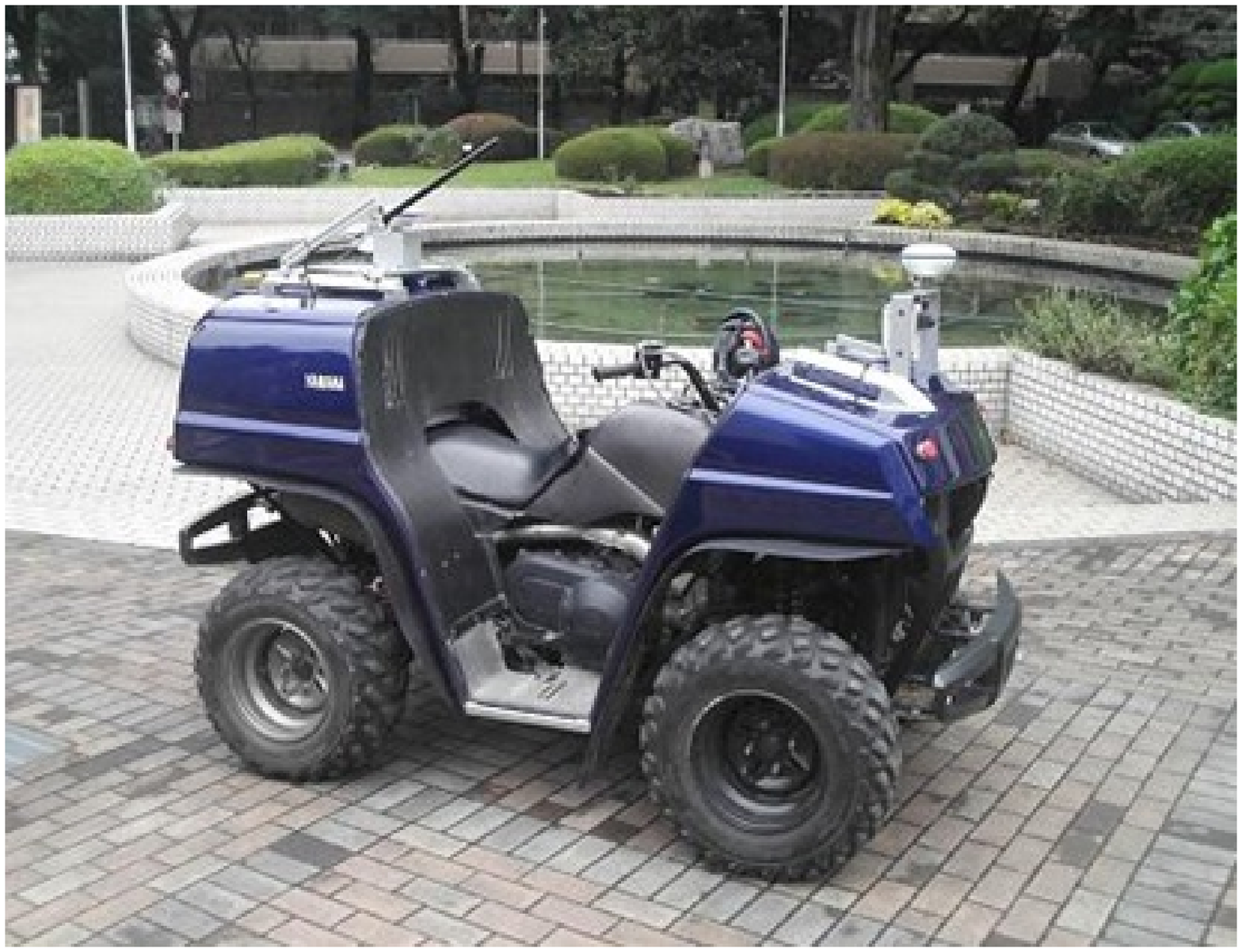}
	\caption{UGV.}
	\label{fig:UGV}
\end{figure}

The outdoor experiment environment is shown in Fig. \ref{fig:course}.
This is a training course of a driving school whose size is about 120[m]
$\times$ 80[m] and the length of the course is about 250[m]. One subject remote controls
the robot three times for each system and
each system is
chosen randomly. in order to cancel the
influence of the order of trials.
The system configuration of this experiment is almost same as
the previous experiment in the section 4.1.
We use a UGV (Unmanned Ground Vehicle) as a mobile robot developed by YAMAHA
Motor Co., Ltd. as shown in Fig. \ref{fig:UGV}.
Maximum transrational velocity of the UGV is 1.0 [m/s] in this experiment.
To control the UGV easily, we use a handle
and a pedal instead of a joystick in the previous experiment.

As we focus on the effectiveness of the zoom function and the additional lines,
the experiments has been carried out without moving objects.
In the experiment, we set the limitation of the communication band with the
assumption of using mobile phone communication. Table \ref{tbl:setting} shows
the parameters of image and data which are sent from the robot to the operator station in each system. These parameters in
each system are set as the best
values to remote control the robot with the limitation of the bandwidth
according to results of preliminary experiments.

\begin{table*}[t]
	\caption{Setting for experiment for zoom function and add lines.}
        \label{tbl:setting}
        \begin{center}
        	\setlength{\tabcolsep}{3pt}
                \footnotesize
		\begin{tabular}{|l|l|l|} \hline
		  & Front Camera & Existing SPIR and Proposed SPIR2 \\ \hline
		 Jpeg quality & 15 & 50 \\ \hline
		 Image size & 640x480[pixel] & 640x480[pixel]\\ \hline
		 Transmission interval of image & 0.7[sec] & 1.4[sec] \\ \hline
		 Transmission delay of image& 1.2[sec]  &  1.9[sec] \\ \hline
		 Transmission interval of data except image& 0.02[sec]  &  0.02[sec]\\ \hline
		 Transmission delay of data except image& 0.5[sec] & 0.5[sec]\\ \hline
		\end{tabular}
	\end{center}
\end{table*}

We record average speeds and position errors as the indicates of
the operator's workload, because if the operator feels complexity of the
teleoperation the speed of the robot will reduce and the position error will
increase.
We define the position error as the sum of the nearest distance between each sampling position of the robot and the center line of the road as the target
trajectory.

\subsection{Results and consideration}

\begin{flushright}
\begin{figure}[t]
 	\centering
 	\includegraphics[width=0.7\hsize]{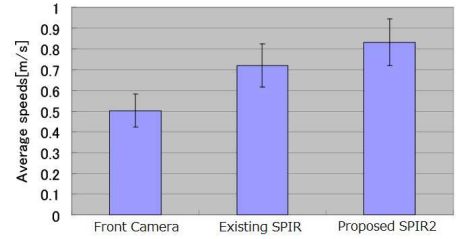}
	\caption{Average speeds.}
	\label{fig:average_speed}
\end{figure}
\end{flushright}

\begin{flushright}
\begin{figure}[t]
 	\centering
 	\includegraphics[width=0.7\hsize]{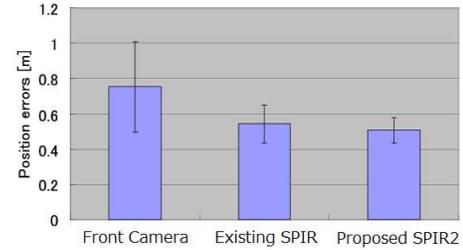}
	\caption{Position errors.}
	\label{fig:running_position_error}
\end{figure}
\end{flushright}

Fig. \ref{fig:average_speed} and Fig. \ref{fig:running_position_error} show the
result of average speeds and average position errors for three methods,
respectively. We evaluate the obtained results with LSD (Least Significant
Difference) test \cite{LSD}.  In the relation of average speeds,
Proposed SPIR2 has significant difference to Existing SPIR, and Existing SPIR
has significant difference to Front Camera.
In the position errors, Front Camera has significant difference
to Existing SPIR and Proposed SPIR2, but there is no significant difference
 between Existing SPIR and Proposed SPIR2.
\par
Because Existing SPIR and Proposed SPIR2 mark high speeds and less errors than
Front Camera, SPIR can apply effectively under the situation of the narrow
communication band. Proposed SPIR2 is better than Existing SPIR at the
average speed. The difference between them is only zoom function and additional
lines. Therefore these proposed improvements are effective for the operator to
teleoperate the robot under the communication limitation.

\par
In the experiment, because each subject teleoperated the robot with three
comparing systems, we use with-in subjects ANOVA.
In order to counterbalance order, each comparing method was randomly taken.
Table \ref{tbl:average_speeds} and Table \ref{tbl:position_errors} shows the
ANOVA table.
In our experimental,
because the F score 34.47 and 7.33 for average speed and position error are also higher than 6.51,
the results are significant at the 0.01 significance level.
\par
Now we apply LSD method for multiple comparison to three comparing systems. In
our experiment, the difference of average is higher than 0.088 or 0.149,
the result is significant at the 0.05 or 0.01 significance level.
We can find the differences over 0.088 among the average speeds of three methods.
In the position errors, the difference between System B and System C is not enough to be significance level, but System A has enough difference to System B and C.
\par
From the results of statistical analysis, the speed of the robot is faster in
the following order: System C (Proposed SPIR2), System B (Existing SPIR) and System A (Front Camera) and the all differences are enough to be 0.05 significance level. On the other hand, the position errors also decrease in same order. However System A (Proposed SPIR2) has enough difference to be significance level, but there is not enough difference between System B (Existing SPIR) and System C (Front Camera).
\par
From these results, we make consideration about each system. With Front Camera,
because the field of view is narrow and the frame rate is decreased, the teleoperation of the robot is very hard. Therefore the trajectory of the robot is meandering shape and the position error is increased. Moreover the operator may teleoperate the robot slowly in order to keep the course.
\par
In the Existing SPIR, the position error is not larger than proposed SPIR2, but
slower. If the background image is not updated, the CG model of the robot on the image is smaller and smaller. However when the image is updated, the size of the CG model is suddenly changed. Because this large change makes the feeling of strangeness to the operator, he cannot input faster command to the robot.
\par
Compared with these systems, Proposed SPIR2 keeps the course and the average speed is very high. The average speed in whole course is 80 

\begin{table}[t]
\caption{ANOVA table (average speeds)}
\label{tbl:average_speeds}
\def\STRUTO{\rule{0mm}{3mm}}%
\def\STRUTU{\rule[-1.5mm]{0mm}{3mm}}%
\begin{center}
\begin{tabular}{lllll}
\hline
\hline
SV & SS & df & MS & F\\
\hline
A     & 0.4453 & 2  & 0.2226 & 32.47 **\\
Sub   & 0.1402 & 7  & 0.0200 & \\
SxA   & 0.0959 & 14 & 0.0068 & \\
\hline
Total & 0.6815 & 23 &        & **p\verb/</.01\\
\end{tabular}
\end{center}
\end{table}

\begin{table}[t]
\caption{ANOVA table (position errors)}
\label{tbl:position_errors}
\def\STRUTO{\rule{0mm}{3mm}}%
\def\STRUTU{\rule[-1.5mm]{0mm}{3mm}}%
\begin{center}
\begin{tabular}{lllll}
\hline
\hline
SV & SS & df & MS & F\\
\hline
A     & 0.2834 & 2  & 0.1417 & 7.33**\\
Sub   & 0.3901 & 7  & 0.0557 & \\
SxA   & 0.2704 & 14 & 0.0193 & \\
\hline
Total & 0.9440 & 23 &        & **p\verb/</.01\\
\end{tabular}
\end{center}
\end{table}

\section{Conclusion}

In this study, we proposed a solution for problem of existing teleoperation
system using past image records (SPIR).
To solve the problem of
existing SPIR that is occurred under narrow communication bandwidth,
zoom function and additional lines are installed in SPIR.
By the outdoor evaluation experiment, we can find that the proposed system is useful under narrow communication band. 
From the experimental results, we find that the proposed SPIR reduces the
operator workloads of teleoperation comparing to existing SPIR.
\par
We would like to extend this system to the multiple-robots
system in the future.

\section*{Acknowledgment}

A part of the results in this research was collaboratively conducted with Yamaha
Motor, Co., Ltd.

\end{document}